\makeatletter\def\graphicscache@inhibit{true}\makeatother
\documentclass[letterpaper, 10 pt, conference]{ieeeconf}

\IEEEoverridecommandlockouts                              %

\usepackage[utf8]{inputenc}

\usepackage[hidelinks,bookmarks=true]{hyperref}
\usepackage[style=ieee,hyperref,natbib=true,backend=biber,firstinits,doi=false,%
     mincitenames=1,maxcitenames=2,maxbibnames=99,sorting=none,terseinits=false,hyperref=true]{biblatex}
\bibliography{references.bib}
\renewbibmacro*{bbx:savehash}{}%
\defbibheading{bibliography}[\bibname]{\section*{References}}

\patchcmd{\bibsetup}{\interlinepenalty=5000}{\interlinepenalty=1000}{}{}

\usepackage{multicol}
\usepackage{multirow}

\usepackage{url}
\usepackage{graphicx}
\usepackage[usenames,dvipsnames]{xcolor}
\usepackage[draft,nomargin,inline]{fixme}
\fxsetup{theme=color,inline=true}

\usepackage{amsmath}
\usepackage{amssymb}
\usepackage{textcomp}
\usepackage{siunitx}

\usepackage{booktabs}
\usepackage{threeparttable}
\usepackage{multirow}

\usepackage{calc}

\usepackage[capitalize]{cleveref}

\usepackage{tikz}
\usetikzlibrary{arrows}
\usetikzlibrary{arrows.meta}
\usetikzlibrary{positioning,calc}
\usetikzlibrary{decorations.pathreplacing}
\usetikzlibrary{decorations.markings}
\usetikzlibrary{fit}
\usetikzlibrary{shapes.callouts}
\usetikzlibrary{shapes.geometric}
\usetikzlibrary{matrix}

\usepackage{pgfplots}
\pgfplotsset{compat=1.9}
\usepgfplotslibrary{groupplots}
\usepgfplotslibrary{units}
\usepgfplotslibrary{statistics}
\usepgfplotslibrary{colorbrewer}
\usepackage{pgfplotstable}

\usepackage{letltxmacro,xcolor,xparse}
\usepackage[export]{adjustbox}

\usepackage{ifthen}

\usepackage{graphicscache}

\usepackage{dblfloatfix}

\tikzset{
  font=\sffamily\footnotesize,
  m/.style={draw, rounded corners, fill=yellow!20, align=center}
}

\pgfplotsset{compat=1.7}

\makeatletter
\tikzdeclarecoordinatesystem{rel}{%
	\tikzset{cs/.cd,x=0pt,y=0pt,#1}%
	\pgfpointlineattime{(\tikz@cs@x)}%
	{\pgfpointanchor{\tikz@pp@name{\tikz@cs@node}}{south west}}%
	{\pgfpointanchor{\tikz@pp@name{\tikz@cs@node}}{south east}}%
	\edef\tikz@cs@x{\the\pgf@x}%
	\pgfpointlineattime{(\tikz@cs@y)}%
	{\pgfpointanchor{\tikz@pp@name{\tikz@cs@node}}{south west}}%
	{\pgfpointanchor{\tikz@pp@name{\tikz@cs@node}}{north west}}%
	\pgfpoint{\tikz@cs@x}{\pgf@y}%
}
\newcommand\currentcoordinate{\the\tikz@lastxsaved,\the\tikz@lastysaved}

\makeatother

\usepackage{blindtext}

\makeatletter
\tikzset{
  border on top/.style={
    append after command={
      node [rectangle,fit=(\tikzlastnode),inner sep=-\pgflinewidth,draw,border on top prevent] {}
    },
  },
  border on top prevent/.code={%
      \let\tikz@after@path\pgfutil@empty%
  },
}
\makeatother

\newsavebox\CBox
\def\textBF#1{\sbox\CBox{#1}\resizebox{\wd\CBox}{\ht\CBox}{\textbf{#1}}}

\title{%
Learning from SAM: Harnessing a Foundation Model for\\
Sim2Real Adaptation by Regularization
}

\author{%
Mayara E. Bonani$^{*1}$, Max Schwarz$^{*1,2,3}$, and Sven Behnke$^{1,2,3}$%
\thanks{* Equal contribution. Contact: \texttt{schwarz@ais.uni-bonn.de}}%
\thanks{$^{1}$Autonomous Intelligent Systems, University of Bonn, Germany}%
\thanks{$^{2}$Lamarr Institute for Machine Learning and AI, Germany}%
\thanks{$^{3}$Center for Robotics, University of Bonn, Germany}%
}

\begin{document}

\maketitle

\begin{abstract}

Domain adaptation is especially important for robotics applications, where
target domain training data is usually scarce and annotations are costly
to obtain.
We present a method for self-supervised domain adaptation for the scenario
where annotated source domain data (e.g. from
synthetic generation) is available, but the target domain data is completely unannotated.
Our method targets the semantic segmentation task and leverages a segmentation
foundation model (Segment Anything Model) to obtain segment information on
unannotated data.
We take inspiration from recent advances in unsupervised local feature learning
and propose an invariance-variance loss 	over the detected segments
for regularizing feature representations in the target domain.
Crucially, this loss structure and network architecture can handle overlapping
segments and oversegmentation as produced by Segment Anything.
We demonstrate the advantage of our method on the challenging YCB-Video and
HomebrewedDB datasets and show that it outperforms prior work and,
on YCB-Video, even a network trained with real annotations.
Additionally, we provide insight through model ablations and show applicability
to a custom robotic application.

\end{abstract}

\section{Introduction}

Domain Adaptation is the art of transferring knowledge learned from one domain
to another domain, usually the application domain.
For roboticists in particular, the very idea of using out-of-domain data is very appealing,
since target domain data is usually not available in quantities that are interesting
for modern deep learning approaches, and target domain \textit{annotations} are even
more costly to obtain.
\begin{tikzpicture}[remember picture,overlay]
  \node[anchor=north,align=center,font=\sffamily\small,yshift=-0.4cm] at (current page.north) {%
  \textbf{Accepted final version.} IEEE International Conference on Automation Science and Engineering (CASE), Los Angeles, USA, August 2025};
\end{tikzpicture}%

Synthetic data is one kind of out-of-domain data, which is particularly cheap
to generate.
It can get close to the target domain in many properties, but still suffers from \textit{domain gap},
the effect that the data distributions do not overlap perfectly and thus a trained model cannot
generalize from training domain to application domain. In the case of synthetic data,
this is more specifically also called the \textit{Sim2Real gap}~\citep{hofer2021sim2real}.

While there is ongoing research on decreasing this gap through improvement of synthetic generators~\citep{schwarz2020stillleben,denninger2020blenderproc} or learning to adapt training data to
fit a target distribution~\citep{imbusch2022synthetic},
we choose to approach the issue from another, orthogonal direction:
Can we regularize the model in a task-specific manner so that it performs more robustly
in the target domain?

Recently, \textit{foundation models}, i.e. very large models trained on internet-scale
datasets, have been released for various tasks.
One of these, the Segment Anything Model (SAM)~\citep{kirillov2023segment} targets segmentation tasks
and demonstrates impressive performance in prompted settings, i.e. where there
is knowledge of the target objects' location.
Still, SAM is not suitable for most robotic applications as-is, since it does not have
any knowledge of the target semantics (e.g. object classes) and is much too
compute-intensive for real-time application.

\begin{figure}
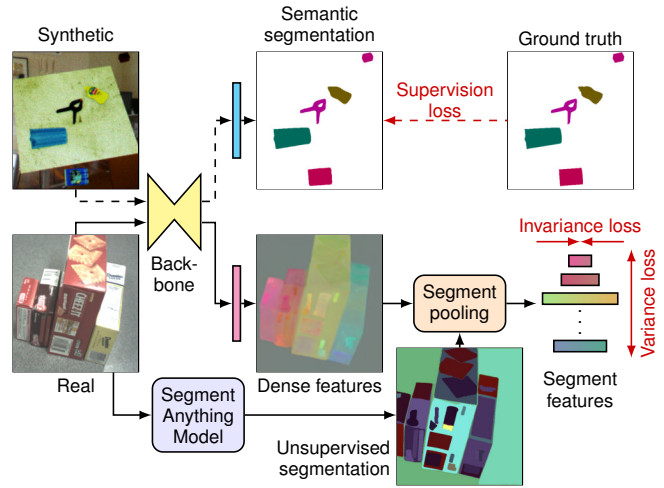

 \newlength\timgheight\setlength\timgheight{1.8cm}
 \centering
\begin{tikzpicture}[
   font=\sffamily\scriptsize,
   layer/.style={
     inner sep=0pt,
     minimum width=0.1cm,
     minimum height=1cm,
     fill=yellow!50,
     draw=black
   },
   segment/.style={
     inner sep=0pt,
     minimum width=0.8cm,
     minimum height=0.15cm,
     fill=yellow!50,
     draw=black
   },
   img/.style={
     draw=black,
     align=center,
     inner sep=0.2pt
   },
   every label/.style={
     inner sep=2pt
   },
   lossn/.style={
     text=red!80!black
   },
   lossp/.style={
     draw=red!80!black
   },
   every path/.style={
     line width=0.6pt
   }
 ]
   \definecolor{seg1a}{rgb}{0.918,0.357,0.608}
   \definecolor{seg1b}{rgb}{0.741,0.482,0.529}
   \definecolor{seg2a}{rgb}{0.780,0.329,0.447}
   \definecolor{seg2b}{rgb}{0.722,0.471,0.443}
   \definecolor{seg3a}{rgb}{0.655,0.898,0.541}
   \definecolor{seg3b}{rgb}{0.894,0.702,0.376}
   \definecolor{seg4a}{rgb}{0.506,0.553,0.714}
   \definecolor{seg4b}{rgb}{0.337,0.659,0.557}

   \node[inner sep=0pt] (backbone) {
     \tikz{
       \draw[black,fill=yellow!50] (0,0.5) -- (0.35,0.1) -- (0.7,0.5)
         -- (0.7,-0.5) -- (0.35,-0.1) -- (0,-0.5) -- cycle;
     }
   };
   \node[below=0cm of backbone,inner sep=2pt,align=center] {Back-\\bone};

   \node[img, anchor=south east,label=above:Synthetic] (synimg) at ($(backbone.west)+(-0.1,0.3)$) {\includegraphics[clip,trim=100 0 100 0,height=\timgheight]{images/teaser/syn_input.png}};
   \node[img, anchor=north east,label=below:Real] (realimg) at ($(backbone.west)+(-0.1,-0.3)$) {\includegraphics[clip,trim=200 0 0 0,height=\timgheight]{images/teaser/real_input_bright.png}};

   \draw[-latex] (realimg.north) -- ++(0,2pt) |- ($(backbone.west)+(0,-4pt)$);
   \draw[-latex,dashed] (synimg.south) -- ++(0,-2pt) |- ($(backbone.west)+(0,4pt)$);

   \coordinate[right=0.4 of backbone] (rightbegin);
   \coordinate[right=0.45 of backbone] (rightstart);
   \node[layer, anchor=west, fill=cyan!50] (seg) at (rightbegin|-synimg) {};
   \node[layer, anchor=west, fill=magenta!50] (proj) at (rightbegin|-realimg) {};
   \draw[-latex,dashed] ($(backbone.east)+(0,4pt)$) -- ++(5pt,0) |- (seg.west);
   \draw[-latex] ($(backbone.east)+(0,-4pt)$) -- ++(5pt,0) |- (proj.west);

   \node[img, right=0.2cm of seg,label={above,align=center:Semantic\\segmentation}] (pred) {\includegraphics[clip,trim=100 0 100 0,height=\timgheight]{images/teaser/syn_pred.png}};
   \draw[-latex] (seg) -- (pred);

   \node[img, right=1.65cm of pred,label={above:Ground truth}] (gt) {\includegraphics[clip,trim=100 0 100 0,height=\timgheight]{images/teaser/syn_label.png}};
   \draw[lossp,latex-,dashed] (pred) -- (gt) node[lossn,midway,align=center,above,inner sep=1pt] {Supervision\\loss};

   \node[img, right=0.2cm of proj,label=below:{Dense features}] (features) {\includegraphics[clip,trim=200 0 0 0,height=\timgheight]{images/teaser/real_features.png}};
   \draw[-latex] (proj) -- (features);

   \node[right=.4cm of features,draw=black,rounded corners,fill=orange!20,align=center] (pool) {Segment\\pooling};

   \node[below=.2cm of pool,img] (samout) {\includegraphics[clip,trim=200 0 0 0,height=\timgheight]{images/teaser/sam_colorized.png}};
   \node[align=right,anchor=south east,inner sep=2pt] at (samout.south west) {Unsupervised\\segmentation};

   \coordinate[right=-0.05 of backbone] (rightbb);
   \node[below=2.2cm of rightbb, rounded corners, draw=black, fill=blue!10,align=center] (sam) {Segment\\Anything\\Model};
   \draw[-latex] ($(realimg.south)+(0.5,0)$) |- (sam);
   \draw[-latex] (sam) -- (sam-|samout.west);
   \draw[-latex] (samout) -- (pool);

   \matrix (segments) [right=.4cm of pool,inner sep=0pt,outer sep=2pt,matrix of nodes, nodes in empty cells, row sep=.08cm] {
     |[segment,minimum width=0.3cm,left color=seg1a,right color=seg1b]| \\
     |[segment,minimum width=0.5cm,left color=seg2a,right color=seg2b]|\\
     |[segment,minimum width=1cm,  left color=seg3a,right color=seg3b]|\\
     |[text height=.3cm]| $\vdots$\\
     |[segment,minimum width=0.7cm,left color=seg4a,right color=seg4b]|\\
   };
   \node[below=5pt of segments, inner sep=0pt, align=center] {Segment\\features};

   \draw[-latex] (features) -- (pool);
   \draw[-latex] (pool) -- (segments);

   \coordinate[right=.1cm of segments] (regloss);
   \draw[lossp,latex-latex] (segments.north east-|regloss) -- (segments.south east-|regloss);
   \node[lossn,rotate=90,anchor=north,align=center] at (regloss) {Variance loss};

   \coordinate[above=.1cm of segments] (invloss);
   \draw[lossp,-latex] (segments.north west|-invloss) -- (invloss);
   \draw[lossp,-latex] (segments.north east|-invloss) -- (invloss);
   \node[lossn,above=0cm of invloss,align=center] {Invariance loss};
\end{tikzpicture}%
 
 \vspace{20pt}
 \caption{Hybrid learning from annotated synthetic data (top path, dashed) and unannotated real data (bottom path, solid).
 For synthetic data, ground truth is available. For real data, projected dense features are aggregated using
 segment information obtained from Segment Anything Model (SAM). An invariance loss forces feature vectors of the same
 segment closer together, while a variance loss spreads segment means apart.
 Both branches of the network train the backbone and thus benefit from each other.
 }
 \label{fig:teaser}
\end{figure}

In this work, we explore a way of distilling the general ``objectness'' knowledge learned by SAM
to achieve robust domain adaptation of semantic segmentation networks. In particular,
we use SAM to generate a regularization signal for the task model---derived from unlabeled target domain data
by pooling dense features into detected segments (see \cref{fig:teaser}).
We then use an invariance-variance loss scheme inspired by recent advancements in
self-supervised feature learning~\citep{bardes2022vicreg,bardes2022vicregl} to
regularize the learned features on unannotated real data.

Our method significantly outperforms a prior Sim2Real method~\citep{imbusch2022synthetic}
and beats even models trained with real labels.

\noindent Our contributions include:
\begin{enumerate}
 \item a segment pooling scheme which uses pre-extracted SAM segments to aggregate
       features,
 \item a loss formulation that allows self-supervised learning on a combination of
       annotated synthetic data and unannotated real data,
 \item thorough evaluation on the challenging YCB-Video and HomebrewedDB datasets
       and two different synthetic data generators,
 \item ablations showing the influence of hyperparameters, and
 \item quick application to a custom robotic problem setting.
\end{enumerate}

\section{Related Work}

Unsupervised domain adaptation is a wide and active research field. While we discuss directly related methods here, we refer to \citet{liu2022deep} for a general overview.

\noindent$\circ$\,\textit{Synthetic Data.}
\Citet{denninger2020blenderproc} propose BlenderProc, focused on the realism of generated images, employing offline rendering techniques such as path tracing using the well-known Blender software.
\Citet{schwarz2020stillleben} introduce the Stillleben framework, which generates training data for perception tasks such as semantic segmentation, object detection, and pose estimation. The images provided by Stillleben are obtained through physics simulation of object meshes and rasterization, allowing the objects to have a randomized appearance and material parameters in addition to noise and transformations that simulate the camera sensors in a scene.
In contrast to BlenderProc, Stillleben can be used online due to its fast GPU-based rendering pipeline.
We evaluate our method using both BlenderProc and Stillleben in this work.

\noindent$\circ$\,\textit{Unsupervised Learning.}
If only unlabeled data is available, one can also turn to unsupervised techniques for learning robust features.
Joint embedding architectures represent one of the main streams in recent works.
\Citet{bardes2022vicreg} proposed  a self-supervised method for training such architectures, called Variance-Invariance-Covariance Regularization (VICReg). %
First, the variance regularization term is defined as a hinge function on the standard deviation of the embeddings along the batch dimension and above a given threshold. It thus forces the embedding vectors of samples within a batch to be different.
Two augmentations of the same sample, in contrast, are forced closer together by the invariance term. Finally, the covariance term encourages decorrelated feature dimensions. Overall, the authors show that their regularization stabilizes training and leads to robust learned features.

The VICReg approach is not directly suitable for image segmentation because it removes spatial information to satisfy the invariance criterion.
To combat this, the VICRegL method by \citet{bardes2022vicregl} adds special consideration of local features by
matching feature vectors pooled from nearby regions in the original image or having a small distance in embedding space.
Our loss formulation is inspired by the VICReg criterion, but in contrast to VICRegL, our method leverages external \mbox{(over-)segmentation} information generated by SAM to derive
intra-frame correspondences. Another difference is that we employ a modern segmentation backbone architecture which
is able to generate features in a much higher resolution.

Perhaps most related to our approach, \Citet{henaff2021efficient} group local feature vectors according to a simple unsupervised pre-segmentation and apply a contrastive objective to each object-level feature separately. Their contrastive objective maximizes the similarity across views of local features, which represent the same object, while minimizing similarity for local features from different objects.
In contrast to their work, which focuses on self-supervised pretraining, our method makes use of the available supervision on synthetic data during training.

\noindent$\circ$\,\textit{Domain Adaptation.}
Although synthetic data is a viable solution to the annotation bottleneck problem,
it suffers from the domain gap between synthetic and real data, i.e. the discrepancy
between the statistical properties obtained by the synthetic data distribution and
the real data distribution. To overcome the performance degradation when the model is trained using real images,  \citet{imbusch2022synthetic} suggested a multi-step learning-based approach to perform synthetic-to-real domain adaptation and consequently minimize the domain gap.
Synthetic images are fed into a domain adaptation network, which generates images more similar to the real dataset while preserving annotations.
This domain adaptation approach is based on
Contrastive Unpaired Translation (CUT)~\citep{park2020contrastive}, which is an
image-to-image translation technique aimed at preserving
the image content while adapting the appearance to a target domain.
In contrast to this method, we do not adapt training images, but instead perform the adaptation while training the task network by regularizing its performance on unannotated data.

\noindent$\circ$\,\textit{Segment Anything Model (SAM).}
Foundation models (such as BERT~\citep{devlin2019bert}, RoBERTa~\citep{liu2019roberta}, and GPT-4~\citep{2023arXiv230308774O}) enable powerful generalization for tasks and data distributions beyond those seen during training. \citet{kirillov2023segment} focused on building a foundation model for image segmentation. 
The promptable segmentation task is to return a valid segmentation mask given any segmentation prompt, such as  a set of foreground and background points or a rough bounding box specifying what to segment in an image. The Segment Anything Model (SAM) can adapt to diverse segmentation tasks.
Importantly for a foundation model, the authors collected the Segment Anything Dataset, SA-1B, which consists of 11 million diverse, high-resolution, licensed, and privacy protecting images and 1.1 billion high-quality segmentation masks. The dataset was collected through model-in-the-loop dataset annotation, leveraging weaker versions of SAM to annotate more training data for later stronger versions.
The trained SAM model can segment unfamiliar objects and images without requiring any training or annotations and can be used for different tasks, as long as a prompt is available.
Additionally, it was designed to be able to naturally handle ambiguity, such as a prompt meaning smaller (e.g. shirt) or bigger (e.g. person) scene entities.
Therefore, its output predicts multiple masks for a single prompt, resulting in segments that  intersect each other.
SAM cannot be applied directly for usual semantic segmentation tasks, though, because it does not produce semantic information.

\Citet{kim2024garfield} proposed a method for utilizing SAM for hierarchical decomposition of scenes.
A very similar contrastive loss structure is used to ensure parts that are grouped by SAM end up with similar features.
Notably, the authors introduce a multi-scale architecture to extend the range of the built grouping hierarchy.
The method is applied to 3D neural radiance fields, focusing on high-fidelity offline decomposition of pre-recorded 3D scenes.
In contrast, our method is applied as regularization during training of a task network, which is later capable of real-time operation on novel scenes,
and, through synthetic supervision, produces semantics.

\section{Method}

We will now describe our method in detail. \cref{fig:teaser} shows the general information flow and \cref{fig:architecture} zooms into the trained network.

\subsection{SAM Preprocessing}
\label{sec:sam_preproc}

\begin{figure}
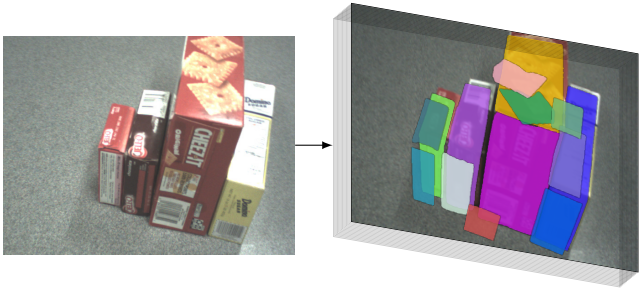

 \setlength\timgheight{2.9cm}
 \centering
 \begin{tikzpicture}
  \node[inner sep=0pt] (input) {\includegraphics[height=\timgheight]{images/teaser/real_input_bright.png}};
 
  \node[right=0.5 of input,inner sep=0] (stack) {\tikz{
  \begin{scope}[
     cm={0.9848077530,-0.1736481777,0,1,(0,0)},z={(-0.1736481777,-0.1736481777)},
     transform shape,
     every node/.style={draw=black,opacity=0.8,inner sep=0pt,draw opacity=0.2,line width=0.2pt}
   ]
   \node[opacity=1] at (0, 0, 0.2) {\includegraphics[height=\timgheight]{images/teaser/sam_input.png}};
   \node at (0, 0, 0.3) {\includegraphics[height=\timgheight]{images/teaser/sam003.png}};
   \node at (0, 0, 0.4) {\includegraphics[height=\timgheight]{images/teaser/sam004.png}};
   \node at (0, 0, 0.5) {\includegraphics[height=\timgheight]{images/teaser/sam005.png}};
   \node at (0, 0, 0.6) {\includegraphics[height=\timgheight]{images/teaser/sam006.png}};
   \node at (0, 0, 0.7) {\includegraphics[height=\timgheight]{images/teaser/sam007.png}};
   \node at (0, 0, 0.8) {\includegraphics[height=\timgheight]{images/teaser/sam008.png}};
   \node at (0, 0, 0.9) {\includegraphics[height=\timgheight]{images/teaser/sam009.png}};
   \node at (0, 0, 1.0) {\includegraphics[height=\timgheight]{images/teaser/sam010.png}};
   \node at (0, 0, 1.1) {\includegraphics[height=\timgheight]{images/teaser/sam011.png}};
   \node at (0, 0, 1.2) {\includegraphics[height=\timgheight]{images/teaser/sam012.png}};
   \node at (0, 0, 1.3) {\includegraphics[height=\timgheight]{images/teaser/sam013.png}};
   \node at (0, 0, 1.4) {\includegraphics[height=\timgheight]{images/teaser/sam014.png}};
   \node at (0, 0, 1.5) {\includegraphics[height=\timgheight]{images/teaser/sam015.png}};
   \node at (0, 0, 1.6) {\includegraphics[height=\timgheight]{images/teaser/sam016.png}};
  \end{scope}
  }};
  
  \draw[-latex] (input) -- (stack);

 \end{tikzpicture}
 \caption{Segments predicted by SAM in its ``segment everything'' mode, shown stacked in 3D on top of the reference image.
 Note how the segments overlap and over-segment the image.
 We only show a small number of segments for clarity.
 }
 \label{fig:samout}
\end{figure}

The Segment Anything Model (SAM, \citep{kirillov2023segment}) allows zero-shot segmentation given a point, mask, or even textual prompt.
It has learned a general objectness concept which allows it to predict the most likely segment belonging to a prompt.
We use the default ``automatic'' mask prediction mode of SAM, which queries the model with a grid of query points to
segment ``everything'' in the image. The default settings are chosen for SAM with a ViT-H backbone and a query grid size
of 24$\times$24 which ensures that objects of typical size are hit by at least one query point.

Because the SAM results only depend on the input image, SAM segmentation can be done in a preprocessing step.
This is most welcome, since SAM execution is compute-intensive.

Because each segment is predicted in isolation, SAM produces an oversegmentation of the image with overlapping masks (see \cref{fig:samout}).
To deal with this effect, we do not directly use the SAM segments to e.g. generate pixel-wise pseudo labels,
which would be difficult due to the overlapping (see \cref{sec:ablations} for an experiment showing this).
Instead, our proposed loss formulation is able to deal with this ambiguity (see Sec.~\ref{sec:SegmentPooling}).

\subsection{Modern Segmentation Backbone}

\begin{figure}
 \setlength\timgheight{2cm}
 \newlength\pimgheight\setlength\pimgheight{.7\timgheight}
 \newlength\pimgwidth\setlength\pimgwidth{1.3333\pimgheight}
 \centering
\begin{tikzpicture}[
   font=\sffamily\scriptsize,
   layer/.style={
     inner sep=0pt,
     minimum width=0.1cm,
     minimum height=1cm,
     fill=yellow!50,
     draw=black
   },
   segment/.style={
     inner sep=0pt,
     minimum width=0.8cm,
     minimum height=0.15cm,
     fill=yellow!50,
     draw=black
   },
   img/.style={
     minimum width=1.33cm,
     minimum height=1cm,
     draw=black,
     align=center,
     inner sep=0.2pt
   },
   every label/.style={
     inner sep=0pt
   },
   m/.style={
    draw=black,fill=yellow!20,
    rounded corners,
    align=center
   },
   lossn/.style={
     text=red!80!black
   },
   lossp/.style={
     draw=red!80!black
   },
   every path/.style={
     line width=0.6pt
   }
 ]
   \begin{scope}[shift={(1.8,-1.23)}]
   \begin{scope}[cm={1,0,0.707,0.707,(0,0)},z={(-1,1.414)},transform shape]
    \node[img,draw=black] (dinput) {\includegraphics[height=\pimgheight]{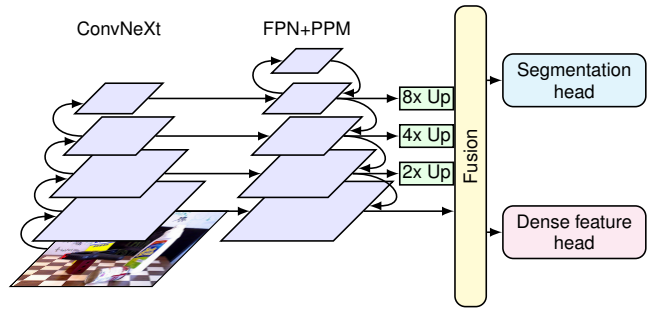}};
    \node[img,fill=blue!10,minimum width=.8\pimgwidth,minimum height=.8\pimgheight] (d0) at (0,0,0.5) {};
    \node[img,fill=blue!10,minimum width=.64\pimgwidth,minimum height=.64\pimgheight] (d1) at (0,0,1) {};
    \node[img,fill=blue!10,minimum width=.512\pimgwidth,minimum height=.512\pimgheight] (d2) at (0,0,1.5) {};
    \node[img,fill=blue!10,minimum width=.4096\pimgwidth,minimum height=.4096\pimgheight] (d3) at (0,0,2) {};

    \coordinate (dinputin) at ($(dinput.west)+(0,-0.1)$);
    \coordinate (d0in) at ($(d0.west)+(0,-0.1)$);
    \coordinate (d1in) at ($(d1.west)+(0,-0.1)$);
    \coordinate (d2in) at ($(d2.west)+(0,-0.1)$);
    \coordinate (d3in) at ($(d3.west)+(0,-0.1)$);

    \node[img,fill=blue!10,minimum width=.8\pimgwidth,minimum height=.8\pimgheight] (u0) at (2.5,0,0.5) {};
    \node[img,fill=blue!10,minimum width=.64\pimgwidth,minimum height=.64\pimgheight] (u1) at (2.5,0,1) {};
    \node[img,fill=blue!10,minimum width=.512\pimgwidth,minimum height=.512\pimgheight] (u2) at (2.5,0,1.5) {};
    \node[img,fill=blue!10,minimum width=.4096\pimgwidth,minimum height=.4096\pimgheight] (u3) at (2.5,0,2) {};
    \node[img,fill=blue!10,minimum width=.32768\pimgwidth,minimum height=.32768\pimgheight] (u4) at (2.5,0,2.5) {};

    \coordinate (u3to4) at ($(u3.west)+(0,0.1)$);

    \coordinate (u0in) at ($(u0.east)+(0,0.1)$);
    \coordinate (u1in) at ($(u1.east)+(0,0.1)$);
    \coordinate (u2in) at ($(u2.east)+(0,0.1)$);
    \coordinate (u3in) at ($(u3.east)+(0,0.1)$);
   \end{scope}
   \end{scope}

   \draw[-latex] (dinput.west) to [out=180,in=180,looseness=3] (d0in);
   \draw[-latex]     (d0.west) to [out=180,in=180,looseness=3] (d1in);
   \draw[-latex]     (d1.west) to [out=180,in=180,looseness=3] (d2in);
   \draw[-latex]     (d2.west) to [out=180,in=180,looseness=3] (d3in);

   \draw[-latex] (d0.east) -- (u0.west);
   \draw[-latex] (d1.east) -- (u1.west);
   \draw[-latex] (d2.east) -- (u2.west);
   \draw[-latex] (d3.east) -- (u3.west);

   \draw[-latex] (u3to4) to [out=180,in=180,looseness=3] (u4.west);

   \draw[-latex]     (u4.east) to [out=0,in=0,looseness=2] (u3in);
   \draw[-latex]     (u3.east) to [out=0,in=0,looseness=3] (u2in);
   \draw[-latex]     (u2.east) to [out=0,in=0,looseness=3] (u1in);
   \draw[-latex]     (u1.east) to [out=0,in=0,looseness=3] (u0in);

   \coordinate (belowpyramid) at (0,1.7);
   \node at (belowpyramid-|d0) {ConvNeXt};
   \node at (belowpyramid-|u0) {FPN+PPM};

   \node[m,anchor=west,minimum height=4cm] (fusion) at ($(0,0-|u0.east)+(1.2,0)$) {\rotatebox{90}{Fusion}};
   \draw[-latex] (u0.east) -- (u0.east-|fusion.west);

   \node[inner sep=1pt,fill=green!10,anchor=east,draw=black] at (u1.east-|fusion.west) (up1) {2x Up};
   \draw[-latex] (u1.east) -- (up1);

   \node[inner sep=1pt,fill=green!10,anchor=east,draw=black] at (u2.east-|fusion.west) (up2) {4x Up};
   \draw[-latex] (u2.east) -- (up2);

   \node[inner sep=1pt,fill=green!10,anchor=east,draw=black] at (u3.east-|fusion.west) (up3) {8x Up};
   \draw[-latex] (u3.east) -- (up3);

   \node[m,anchor=north west,minimum width=1.9cm,fill=magenta!10] (projhead) at ($(fusion.east)+(0.2,-0.65)$) {Dense feature\\head};
   \draw[-latex] (fusion.east|-projhead) -- (projhead);

   \node[m,anchor=south west,minimum width=1.9cm,fill=cyan!10] (seghead) at ($(fusion.east)+(0.2,0.65)$) {Segmentation\\head};
   \draw[-latex] (fusion.east|-seghead) -- (seghead);
 \end{tikzpicture}
  \caption{
   Detailed architecture of the backbone with the segmentation head and the dense feature head.
   ``Up'' denotes a bilinear upsampling layer.
 }
 \label{fig:architecture}
\end{figure}

The backbone of our network follows a standard architecture in recent segmentation
works~\citep{xiao2018unified,kirillov2019panoptic,liu2022convnet}:
We combine a strong ConvNeXt-L~\citep{liu2022convnet} pretrained on ImageNet classification
with a Feature Pyramid Network (FPN)~\citep{lin2017feature}, which propagates highly semantic features
from higher, low-resolution layers back to the lower, high-resolution layers (see \cref{fig:architecture}).
Inspired by \citet{xiao2018unified}, we also add an additional Pyramid Pooling Module (PPM)~\citep{zhao2017pyramid} on top of the feature pyramid,
which improves global context by even further spatially aggregating features.
A single fusion layer than produces a single feature map with 512 channels and high resolution.
The resulting structure is similar to the one evaluated in \citep{liu2022convnet} for
segmentation tasks. Congruent with ConvNeXt, all our network layers use GELU activations.

\subsection{Segmentation Head}

A single convolutional layer with kernel size 1$\times$1 computes
the semantic segmentation result (i.e. class logits). On synthetic images, where annotations are available,
this layer and the backbone are trained using the standard cross entropy loss $\mathcal{L}_\text{Sup}$.
For images without annotations, this output provides a compatible semantic segmentation but due to the lack of ground truth, it cannot be used for training.

\subsection{Dense Feature Head}

Inspired by VICReg~\citep{bardes2022vicreg} and VICRegL~\citep{bardes2022vicregl},
we add a projection block for the self-supervised feature learning on real images.
This projection block consists of two layers that bring the feature dimension first
down to 256 and then to~$D$.
In our experiments we choose $D=3$, which has the advantage of being easily visualizable
as RGB colors. Low dimensionality in contrastively learned features is a common choice~\citep{florence2018dense}
and indeed, higher $D$ does not lead to better performance (see \cref{sec:ablations}).

The fact that the self-supervised feature learning uses a different projector than
the semantic segmentation provides the necessary decoupling: Since SAM oversegments
the image, it is important that the segmentation head can ignore some fine-grained
differences visible in the dense feature output (see the exemplary SAM output in
\cref{fig:samout}).

\subsection{Segment Pooling}
\label{sec:SegmentPooling}

\citet{henaff2021efficient} pool feature vectors coming from the same region (as predicted by an EMA teacher model) in the mask and contrast the resulting vectors between each other with a contrastive loss function. This allows feature vectors spatially far away in the original image to be pooled together if they belong to the same object.
Following a similar idea, we use predicted SAM segments to pool features. Since SAM predictions are available for both synthetic and real data, this
allows regularization in both domains. Furthermore, the regularization will be stable, since SAM predictions are constant during training.

After upsampling the output of the dense feature head to the image resolution,
our pooling block computes the mean $\mu_i \in \mathbb{R}^D$ and the unnormalized variance vector $v_i \in \mathbb{R}^D$ for each segment $i \in \{0,\dots,N\}$:
\begin{alignat}{2}
 \mu_i &= \frac{1}{|S_i|} \sum_{p \in S_i} z_p \text{, and} \\
 v_i &= \sum_{p \in S_i} (z_p - \mu_i)^{\odot 2},
\end{alignat}
where $S_i$ is the set of pixels belonging to segment $i$, $z_p$ is the feature vector located at $p$,
and $(\cdot)^{\odot 2}$ denotes the element-wise squaring operation (Hadamard power).
The variance is not normalized by pixel count, since this would give equal importance to segments of any size, resulting
in an undesired focus on smaller segments.
Instead, we give each pixel of each segment equal weighting (see below).

\textit{$\circ$\,Invariance Loss.}
The first information we can obtain from SAM is that pixels from the same segment are likely to belong to the
same object. Thus, we strive to keep segment variance low. Therefore, we define the invariance loss
\begin{equation}\label{eq:invar}
  \mathcal{L}_\text{Inv}  = \frac{\frac{1}{D} \sum_{i=1}^N ||v_{i}||_1 }{\sum_{i=1}^N |S_i|}\,.
\end{equation}

\textit{$\circ$\,Variance Loss.}
The second insight is that two different SAM segments are likely from different objects. We thus expect
a minimum margin $\beta$ between means, leading to the variance loss
\begin{equation}\label{eq:var}
  \mathcal{L}_\text{Var} = \frac{2}{N(N-1)} \sum_{i\neq j}\max(0, \beta-||\mu_i-\mu_j||_2)\,,
\end{equation}
where we choose $\beta = 0.5$ as in \citep{bardes2022vicreg}.

\Citet{bardes2022vicreg} also propose a covariance loss which helps to keep individual feature dimensions decorrelated.
In our experiments, we found that this component is unnecessary in our case, most likely because of the presence of supervision
on synthetic images, which yields backbone outputs (before the heads) that correspond to object classes.
However, both invariance and variance losses are necessary---training with invariance loss only leads to uniform predictions,
while variance loss alone leads to quasi-random predictions.

The total loss in the self-supervised case on real images is thus
\begin{equation}
 \mathcal{L}_\text{real} = \alpha \left( \mathcal{L}_\text{Inv} + \mathcal{L}_\text{Var} \right),
\end{equation}
where $\alpha = 0.05$ is used to scale the loss in relation to the supervised loss on synthetic images,
bringing them approximately to the same magnitude on the YCB-Video dataset.

\begin{figure*}
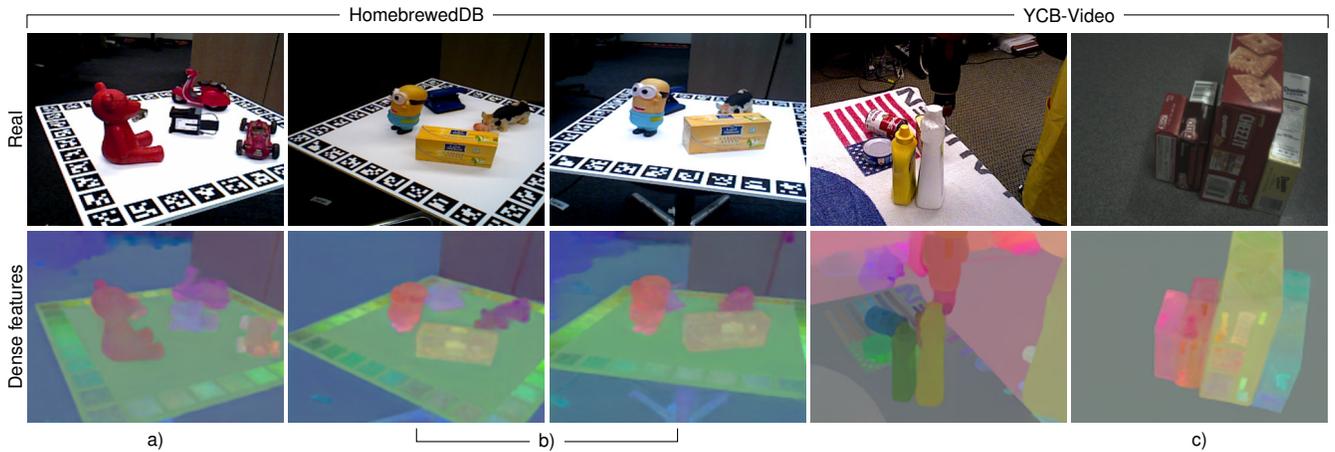

 \centering
 \newlength\fh\setlength\fh{2.55cm}
 \begin{tikzpicture}[font=\sffamily\scriptsize]
  \matrix[matrix of nodes,nodes={inner sep=0pt,anchor=center},row sep=2pt,column sep=2pt] {
    \rotatebox{90}{Real} &
    |(h1)| \includegraphics[height=\fh]{images/features/feat1real.png} &
    \includegraphics[height=\fh]{images/features/feat2real.png} &
    |(h2)| \includegraphics[height=\fh]{images/features/feat3real.png} &
    |(y1)| \includegraphics[height=\fh]{images/features/feat4real.png} &
    |(y2)| \includegraphics[height=\fh]{images/features/feat5real.png} \\
    \rotatebox{90}{Dense features} &
    |(a)|  \includegraphics[height=\fh]{images/features/feat1.png} &
    |(b1)| \includegraphics[height=\fh]{images/features/feat2.png} &
    |(b2)| \includegraphics[height=\fh]{images/features/feat3.png} &
           \includegraphics[height=\fh]{images/features/feat4.png} &
    |(c)|  \includegraphics[height=\fh]{images/features/feat5.png} \\
  };
  \draw (h1.north west) ++(0,2pt) -- ++(0,5pt) -- (\currentcoordinate-|h2.north east) node[midway,fill=white] {HomebrewedDB} -- ++(0,-5pt);
  \draw (y1.north west) ++(0,2pt) -- ++(0,5pt) -- (\currentcoordinate-|y2.north east) node[midway,fill=white] {YCB-Video} -- ++(0,-5pt);
  
  \node[below=0cm of a] {a)};
  \draw (b1.south) ++(0,-2pt) -- ++(0,-5pt) -- (\currentcoordinate-|b2) node[midway,fill=white] {b)} -- ++(0,5pt);
  \node[below=0cm of c] {c)};
 \end{tikzpicture}
 \caption{Learned dense features normalized and visualized as RGB colors.
 The upper row shows input images from HomebrewedDB (left) and YCB-Video (right).
 The bottom row shows features learned by the network in self-supervised fashion.
 Note that a) features correspond well to objects, b) features are stable under camera motion, and c) SAM oversegmentation
 leads to object parts such as text receiving slightly different but related locations in feature space.}
 \label{fig:features}
\end{figure*}

\textit{What happens if SAM oversegments?}
In this case, SAM produces overlapping segments with segments corresponding to the entire object and segments for individual parts~\citep{kirillov2023segment}.
Here,
variance and invariance loss will both be applied to the same pixel pairs. This balanced regularization yields similar, but distinct features for objects and their parts, as discussed in \cref{sec:results} and shown in \cref{fig:features}, and thus poses no problem, since objects can still clearly be differentiated.
In general, we emphasize that our primary goal is not to learn dense features corresponding to target classes, but to regularize the segmentation network.
The segmentation head can, by learning from its synthetic supervision, decide to ignore adverse effects that arise from SAM regularization,
as long as the undersegmentation or oversegmentation behavior of SAM also occurs in synthetic data.

We also note that it would be possible to further refine or filter SAM masks by using the model trained on synthetic data as a teacher.
However, our regularization approach is designed to help especially in the cases where the unregularized method would make
segmentation mistakes, and we argue that it does not help to commit to these mistakes by focusing on SAM masks that undersegment or oversegment the
target object. Indeed, \cref{sec:ablations} shows that ``intelligently'' generating pseudo-labels in such a manner does not improve performance.

\section{Evaluation}\label{sec:eval}

\begin{figure}
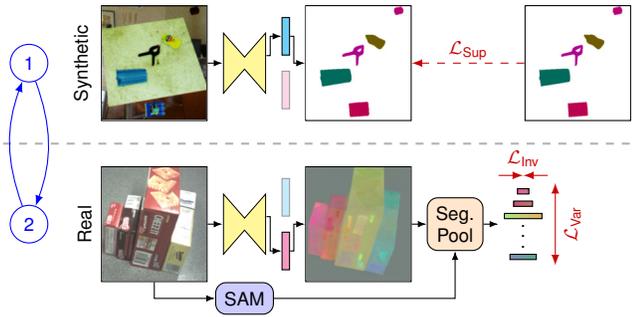

 \setlength\timgheight{1.5cm}
 \centering
\begin{tikzpicture}[
   font=\sffamily\scriptsize,
   img/.style={draw=black,inner sep=0.4pt},
   layer/.style={
     inner sep=0pt,
     minimum width=0.1cm,
     minimum height=.5cm,
     fill=yellow!50,
     draw=black
   },
   segment/.style={
     inner sep=0pt,
     minimum width=0.8cm,
     minimum height=0.07cm,
     fill=yellow!50,
     draw=black
   },
   a/.style={-{Latex[scale=0.7]}},
   lossn/.style={
     text=red!80!black
   },
   lossp/.style={
     draw=red!80!black
   },
   stepc/.style={
     circle,minimum size=0.5cm,draw=black,inner sep=0pt,color=blue
   }
  ]
  \definecolor{seg1a}{rgb}{0.918,0.357,0.608}
   \definecolor{seg1b}{rgb}{0.741,0.482,0.529}
   \definecolor{seg2a}{rgb}{0.780,0.329,0.447}
   \definecolor{seg2b}{rgb}{0.722,0.471,0.443}
   \definecolor{seg3a}{rgb}{0.655,0.898,0.541}
   \definecolor{seg3b}{rgb}{0.894,0.702,0.376}
   \definecolor{seg4a}{rgb}{0.506,0.553,0.714}
   \definecolor{seg4b}{rgb}{0.337,0.659,0.557}

  \node[img,anchor=south west] (input) {\includegraphics[clip,trim=100 0 100 0,height=\timgheight]{images/teaser/syn_input.png}};

  \node[left=0 of input] {\rotatebox{90}{Synthetic}};

  \node[left=0.7 of input,stepc] (step1) {1};

  \node[right=0.2 of input,inner sep=0pt,scale=0.8] (backbone) {
     \tikz{
       \draw[black,fill=yellow!50] (0,0.5) -- (0.35,0.1) -- (0.7,0.5)
         -- (0.7,-0.5) -- (0.35,-0.1) -- (0,-0.5) -- cycle;
     }
   };
   \node[layer,anchor=south west,fill=cyan!50] (seg) at ($(backbone.east)+(0.2,0.1)$) {};
   \node[layer,anchor=north west,opacity=0.4,fill=magenta!50] (proj) at ($(backbone.east)+(0.2,-0.1)$) {};

   \coordinate (rightstart) at ($(seg.east)+(0.2,0)$);

   \node[img,anchor=west] (pred) at (rightstart|-backbone) {%
     \includegraphics[clip,trim=100 0 100 0,height=\timgheight]{images/teaser/syn_pred.png}
   };

   \node[img,right=1.5 of pred] (gt) {%
     \includegraphics[clip,trim=100 0 100 0,height=\timgheight]{images/teaser/syn_label.png}
   };

   \draw[a] (input) -- (backbone);
   \draw[a] ($(backbone.east)+(0,0.1)$) -- ++(0.05,0) |- (seg);
   \draw[a] (seg.east) -- ++(0.05,0) |- (pred);
   \draw[lossp,latex-,dashed] (pred) -- (gt) node[lossn,midway,align=center,above,inner sep=1pt] {$\mathcal{L}_\text{Sup}$};

   \draw[thick,dashed,draw=black!30] (-1.3,-0.3) -- (7.15,-0.3);

   \node[img,below=0.6 of input] (rinput) {\includegraphics[clip,trim=200 0 0 0,height=\timgheight]{images/teaser/real_input_bright.png}};
   \node[left=0 of rinput] {\rotatebox{90}{Real}};
   \node[left=0.7 of rinput,stepc] (step2) {2};

  \node[right=0.2 of rinput,inner sep=0pt,scale=0.8] (backbone) {
     \tikz{
       \draw[black,fill=yellow!50] (0,0.5) -- (0.35,0.1) -- (0.7,0.5)
         -- (0.7,-0.5) -- (0.35,-0.1) -- (0,-0.5) -- cycle;
     }
   };
   \node[layer,anchor=south west,opacity=0.4,fill=cyan!50] (seg) at ($(backbone.east)+(0.2,0.1)$) {};
   \node[layer,anchor=north west,fill=magenta!50] (proj) at ($(backbone.east)+(0.2,-0.1)$) {};

   \coordinate (rightstart) at ($(seg.east)+(0.2,0)$);

   \node[img,anchor=west] (features) at (rightstart|-backbone) {%
     \includegraphics[clip,trim=200 0 0 0,height=\timgheight]{images/teaser/real_features.png}
   };

   \node[right=.2cm of features,draw=black,rounded corners,fill=orange!20,align=center] (pool) {Seg.\\Pool};

   \matrix (segments) [right=.2cm of pool,inner sep=0pt,outer sep=2pt,matrix of nodes, nodes in empty cells, row sep=.08cm] {
     |[segment,minimum width=0.15cm,left color=seg1a,right color=seg1b]| \\
     |[segment,minimum width=0.25cm,left color=seg2a,right color=seg2b]|\\
     |[segment,minimum width=0.5cm,  left color=seg3a,right color=seg3b]|\\
     |[text height=.3cm]| $\vdots$\\
     |[segment,minimum width=0.35cm,left color=seg4a,right color=seg4b]|\\
   };

   \coordinate[right=.1cm of segments] (regloss);
   \draw[lossp,latex-latex] (segments.north east-|regloss) -- (segments.south east-|regloss);
   \node[lossn,rotate=90,anchor=north,align=center] at (regloss) {$\mathcal{L}_\text{Var}$};

   \coordinate[above=.1cm of segments] (invloss);
   \draw[lossp,-latex] (segments.north west|-invloss) -- (invloss);
   \draw[lossp,-latex] (segments.north east|-invloss) -- (invloss);
   \node[lossn,above=0cm of invloss,align=center] {$\mathcal{L}_\text{Inv}$};

   \node[anchor=north,rounded corners,draw=black,fill=blue!20] (sam) at (rinput.south-|backbone) {SAM};

   \draw[a] (rinput) -- (backbone);
   \draw[a] ($(backbone.east)+(0,0.1)$) -- ++(0.05,0) |- (proj);
   \draw[a] (proj.east) -- ++(0.05,0) |- (features);
   \draw[a] (features) -- (pool);
   \draw[a] (pool) -- (segments);
   \draw[a] (rinput.south) |- (sam);
   \draw[a] (sam) -| (pool);

   \draw[-latex,color=blue] (step1) to[bend left=20] (step2);
   \draw[-latex,color=blue] (step2) to[bend left=20] (step1);
 \end{tikzpicture}
  \caption{Alternating training of the backbone and heads on synthetic (top) and real (bottom) data.}
 \label{fig:alternating}
\end{figure}

\subsection{Training Details}

We follow the training and evaluation protocol of \citet{imbusch2022synthetic}:
Training is conducted over 300 epochs of 1500 frames each with a batch size of 1. Synthetic data is either generated on the fly (so that a frame
is never repeated) or selected randomly from available frames. Accordingly, for real data the 1500 frames are randomly chosen from the entire dataset for each epoch.
Similar to \citep{imbusch2022synthetic}, an exponential moving average (EMA) version of the model with $\alpha_\textrm{EMA}=0.995$ is used for evaluation.
We report the mean intersection over union (mIoU) score on the test dataset, averaged over the last 50 training epochs,
which eliminates any remaining short-term training effects (see \citep{imbusch2022synthetic} for details).

Our full model requires both synthetic and real images. To keep RAM usage low, we follow an alternating training scheme (see \cref{fig:alternating}),
where in each iteration first a synthetic image is processed and then a real image. The supervised and self-supervised
losses are thus only added stochastically over the course of optimization.
Training was conducted on NVIDIA RTX A6000 and A100 GPUs.

\subsection{Datasets}

As in \citep{imbusch2022synthetic}, we evaluate our method on the YCB-Video~\citep{xiang2018posecnn} and HomebrewedDB~\citep{kaskman2019homebreweddb} datasets.
Both datasets are part of the well-known BOP challenge~\citep{sundermeyer2023bop} and contain object
meshes---a prerequisite for generating synthetic data.

\textit{$\circ$\,YCB-Video} contains 21 YCB~\citep{calli2015ycb} objects captured with an \mbox{RGB-D} camera in 92 videos, totaling approx. 134k frames.
The frames exhibit difficult lighting conditions and camera noise.
For synthetic data, we use Stillleben~\citep{schwarz2020stillleben}, which generates
physically plausible arrangements and renderings on the fly during training.

\textit{$\circ$\,HomebrewedDB} is a slightly smaller dataset with 13 different scenes. Following \citep{imbusch2022synthetic},
we only use the PrimeSense frames. The ``val'' split is used for training, and the ``test'' split
for evaluation (HomebrewedDB does not contain a ``train'' split with real data).
The semantic segmentation is trained and evaluated only on the BOP test objects.
For synthetic data, we use the BlenderProc4BOP synthetic frames provided by BOP.
\citet{imbusch2022synthetic} used Stillleben data for HomebrewedDB, but since Stillleben is not well
adapted to HomebrewedDB, which contains many occluding background objects, we observed lower performance.
The BlenderProc4BOP images also allow us to demonstrate our method on a second type of synthetic data.

\subsection{Results}\label{sec:results}

\begin{table}
 \centering
 \begin{threeparttable}
  \caption{Results on YCB Datasets}\label{tab:results}
  \begin{tabular}{lrr}
  \toprule
         & \multicolumn{2}{c}{Mean IoU} \\
  \cmidrule (l) {2-3}
  Method \textbackslash ~Dataset

 & YCB-Video~\citep{xiang2018posecnn} & HomebrewedDB~\citep{kaskman2019homebreweddb} \\
  \midrule
  \citet{imbusch2022synthetic} \\
  - real labels           & 0.770 & 0.737 \\
  - synthetic only        & 0.701 & 0.481\tnote{1} \\
  - full                  & 0.763 & 0.558\tnote{1} \\
  \midrule
  Ours\\
  - real labels           & 0.839 & \textBF{0.883} \\
  - synthetic only        & 0.807 & 0.748 \\
  - CUT~\citep{imbusch2022synthetic} only\tnote{2} & 0.814 & - \\
  - full                  & \textBF{0.853} & 0.787\tnote{3} \\
  \bottomrule
 \end{tabular}
 Note: ``real labels'' is a baseline which has access to real supervision.
 \begin{tablenotes}
  \item[1] \citep{imbusch2022synthetic} uses Stillleben~\citep{schwarz2020stillleben} synthetic data, we use BlenderProc4BOP for HomebrewedDB.
  \item[2] Training our backbone on CUT-refined synthetic data.
  \item[3] Model was trained for only 200 epochs.
 \end{tablenotes}
 \end{threeparttable}
\end{table}

\Cref{tab:results} reports semantic segmentation results on YCB-Video and HomebrewedDB.
We compare our results with those of \citet{imbusch2022synthetic}, since they address the same
problem setting.
Similar to their evaluation, is not our intention to beat the overall state-of-the-art in semantic segmentation, but instead show the improvement gained by using our method when applied to a straightforward segmentation model.

As a first observation, our more modern segmentation backbone based on ConvNeXt and FPN+PPM compared to
the older RefineNet~\citep{lin2017refinenet} used in \citep{imbusch2022synthetic} is able to achieve much higher mIoU on the baseline of real annotated data for both datasets.
To analyze the impact of this further, we train an ablation of our backbone with CUT-refined data, like in \citep{imbusch2022synthetic}. This combination achieves similar results as compared to training our backbone directly on synthetic data.
In conclusion, the more modern backbone architecture already provides the robustness that was induced through
data augmentation in \citep{imbusch2022synthetic}.

Furthermore, our method not only outperforms a network purely trained on synthetic data, but also beats
the real baseline on YCB-Video.
On HomebrewedDB, our method provides a decent advantage over synthetic data alone, but does not reach the
performance possible from real annotations. This may be due to the smaller quantity and especially the smaller
diversity of available real data.

\Cref{fig:features} shows the learned dense features qualitatively.
It is immediately evident that the learned features correspond well to the objects to be segmented and
it can be concluded that learning such a representation is highly likely to be helpful for the semantic segmentation task.
The fact that the learned representations remain stable under camera motion is a further indicator that the network
is learning a shared representation for semantic segmentation and dense features---since the invariance-variance loss alone
does not demand such coherence.
Interestingly, the feature visualizations exhibit the over-segmentation done by SAM. Small parts of objects,
such as text and/or markings receive slightly different feature values in order to satisfy the variance
loss. Still, through the decoupled two-head design, this does not impact segmentation performance.

\newlength{\qwi}\setlength\qwi{2.2cm}
\newcommand{\eimg}[1]{\includegraphics[width=\qwi]{images/eval/#1}}
\tikzset{
  eimgmatrix/.style={
    matrix, matrix of nodes, row sep=-.5\pgflinewidth, column sep=-.5\pgflinewidth,inner sep=0pt
  },
  eimg/.style={border on top}
}

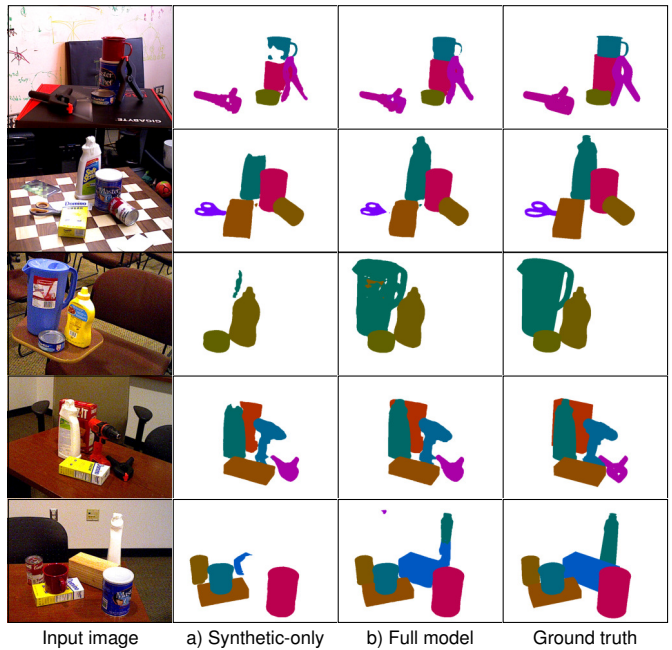
\begin{figure}
 \centering
 \begin{tikzpicture}[font=\sffamily\scriptsize]
  \matrix[eimgmatrix] {
   |[eimg]| \eimg{ycb_syn/real01.png} &
   |[eimg]| \eimg{ycb_syn/pred01.png} &
   |[eimg]| \eimg{ycb_samreg/pred01.png} &
   |[eimg]| \eimg{ycb_samreg/label01.png}
   \\
   |[eimg]| \eimg{ycb_syn/real03.png} &
   |[eimg]| \eimg{ycb_syn/pred03.png} &
   |[eimg]| \eimg{ycb_samreg/pred03.png} &
   |[eimg]| \eimg{ycb_samreg/label03.png}
   \\
   |[eimg]| \eimg{ycb_syn/real04.png} &
   |[eimg]| \eimg{ycb_syn/pred04.png} &
   |[eimg]| \eimg{ycb_samreg/pred04.png} &
   |[eimg]| \eimg{ycb_samreg/label04.png}
   \\
   |[eimg]| \eimg{ycb_syn/real05.png} &
   |[eimg]| \eimg{ycb_syn/pred05.png} &
   |[eimg]| \eimg{ycb_samreg/pred05.png} &
   |[eimg]| \eimg{ycb_samreg/label05.png}
   \\
   |[eimg]| \eimg{ycb_syn/real06.png} &
   |[eimg]| \eimg{ycb_syn/pred06.png} &
   |[eimg]| \eimg{ycb_samreg/pred06.png} &
   |[eimg]| \eimg{ycb_samreg/label06.png}
   \\[3pt]
   Input image & a) Synthetic-only & b) Full model & Ground truth \\
  };
 \end{tikzpicture}
 \caption{Qualitative segmentation results on the YCB-Video test set.
 Compared to a model trained only on synthetic data (a),
 the predictions of the model trained with SAM regularization (b) are generally more complete.}
 \label{fig:segresults}
\end{figure}

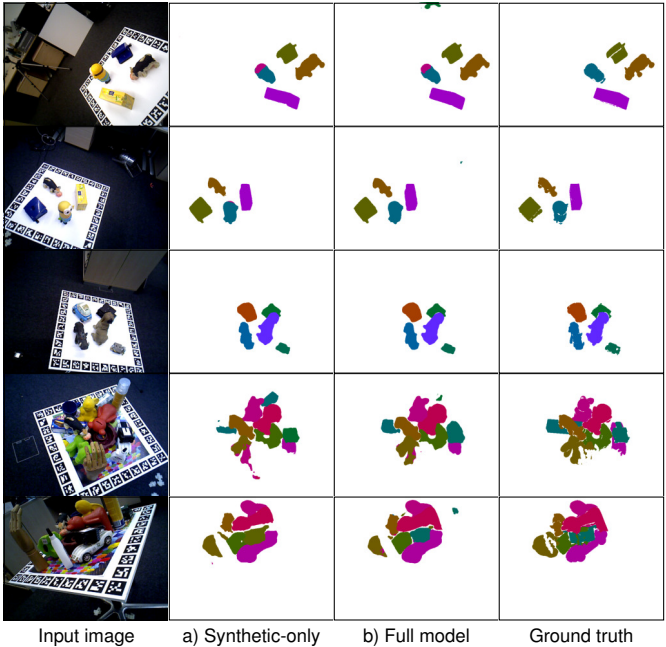
\begin{figure}
 \centering
 \begin{tikzpicture}[font=\sffamily\scriptsize]
  \matrix[eimgmatrix] {
   |[eimg]| \eimg{hddb_syn/real01.png} &
   |[eimg]| \eimg{hddb_syn/pred01.png} &
   |[eimg]| \eimg{hddb_samreg/pred01.png} &
   |[eimg]| \eimg{hddb_samreg/label01.png}
   \\
   |[eimg]| \eimg{hddb_syn/real03.png} &
   |[eimg]| \eimg{hddb_syn/pred03.png} &
   |[eimg]| \eimg{hddb_samreg/pred03.png} &
   |[eimg]| \eimg{hddb_samreg/label03.png}
   \\
   |[eimg]| \eimg{hddb_syn/real05.png} &
   |[eimg]| \eimg{hddb_syn/pred05.png} &
   |[eimg]| \eimg{hddb_samreg/pred05.png} &
   |[eimg]| \eimg{hddb_samreg/label05.png}
   \\
   |[eimg]| \eimg{hddb_syn/real07.png} &
   |[eimg]| \eimg{hddb_syn/pred07.png} &
   |[eimg]| \eimg{hddb_samreg/pred07.png} &
   |[eimg]| \eimg{hddb_samreg/label07.png}
   \\
   |[eimg]| \eimg{hddb_syn/real08.png} &
   |[eimg]| \eimg{hddb_syn/pred08.png} &
   |[eimg]| \eimg{hddb_samreg/pred08.png} &
   |[eimg]| \eimg{hddb_samreg/label08.png}
   \\[3pt]
   Input image & a) Synthetic-only & b) Full model & Ground truth \\
  };
 \end{tikzpicture}
 \caption{Qualitative segmentation results on the HomebrewedDB test set.
 Compared to a model trained only on synthetic data (a),
 the predictions of the model trained with SAM regularization (b) are more complete and show less classification errors.}
 \label{fig:segresults_hd}
\end{figure}

If we analyze the regularizing effect of SAM on the segmentation output, we can see that
using the full pipeline encourages more complete segments (see \cref{fig:segresults,fig:segresults_hd}).
In contrast, a model trained only on synthetic data will often only detect parts of
objects.
In our opinion, this means that the general sense of ``objectness'' learned by SAM has been
successfully distilled and combined with the target domain semantics by our approach.

Executing our model requires only one forward pass and is possible in real-time
on an NVIDIA RTX A6000 GPU (40\,ms per 640$\times$480 frame). In contrast, SAM
with grid prompting takes more than one second to process the same frame.

\subsection{Additional Ablations}
\label{sec:ablations}

We performed additional ablations to investigate the contribution and parameters of
SAM regularization.

\paragraph{Amount of required data}

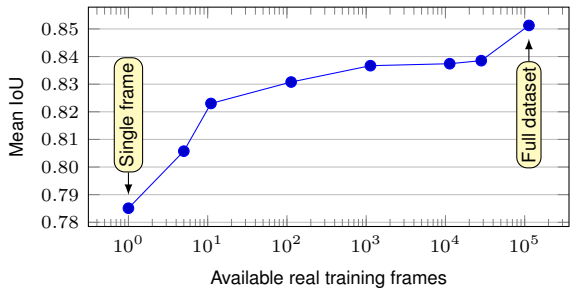
\begin{figure}
  \centering
  \begin{tikzpicture}[font=\sffamily\scriptsize]
    \begin{axis}[
        width=0.9\linewidth, height=4.5cm,
        xmode=log,
        xlabel={Available real training frames},ylabel={Mean IoU},
        ytick distance=0.01,ymajorgrids,
        yticklabel style={/pgf/number format/.cd,fixed,fixed zerofill,precision=2}
      ]
      \addplot+ [
      ] table [x=frames,y=iou] {data/ablation_subset.txt};

      \draw[latex-]
       (axis cs:1,0.785093) ++(0,5pt) -- ++(0,0.3cm)
       node[rotate=90,anchor=west,inner sep=2pt,rounded corners,fill=yellow!30,draw] {Single frame};
      \draw[latex-]
       (axis cs:113000,0.851267) ++(0,-5pt) -- ++(0,-0.3cm)
       node[rotate=90,anchor=east,inner sep=2pt,rounded corners,fill=yellow!30,draw] {Full dataset};

    \end{axis}
  \end{tikzpicture}
  \vspace{-1.5ex}
  \caption{Sensitivity to amount of real (unlabeled) training data. The model was trained on randomly sampled subsets of the YCB-Video training set.
  All models were trained for 200 epochs. The full training set has 113k frames.}
  \label{fig:subset}
\end{figure}

In an application setting, it is desirable to know how many (unlabeled) real frames
have to be captured. We performed training runs with randomly subsampled training sets
on YCB-Video (see \cref{fig:subset}).
While training with a single real frame hurts performance compared with the purely-synthetic
version (compare \cref{tab:results}), gains are observed starting at 10 frames.
This shows that our method is very cheap to apply as it does not require extensive dataset
capture.

\begin{table}
  \centering
  \caption{Additional Ablations and Variants}\label{tab:ablations}
  \begin{threeparttable}
  \begin{tabular}{l|lr}
    \toprule
    Experiment & Variant & Mean IoU \\
    \midrule
    \multirow{2}{*}{Pseudolabels} & Synthetic only     & 0.807 \\
      & SAM pseudo-labels   & 0.805 \\
    & SAM regularization & \textBF{0.853} \\
    \midrule
    \multirow{2}{*}{Domain gap}& Wider gap (syn only) & 0.106 \\
     & Wider gap (full)     & \textBF{0.771} \\
    \midrule
    \multirow{2}{*}{Loss scaling}& $\alpha=0.025$         & 0.832 \\
     &$\alpha=0.05$ (chosen) & \textBF{0.853} \\
     & $\alpha=0.1$           & 0.852 \\
    \midrule
    \multirow{2}{*}{Feature dimension}& $D=12$                 & 0.845 \\
     & $D=6$                  & 0.850 \\
    & $D=3$ (chosen)         & \textBF{0.853} \\
    & $D=2$                  & 0.847 \\
    & $D=1$                  & 0.847 \\
    \midrule
    \multirow{2}{*}{Robotic application}& Synthetic only & 0.704 \\
     & Full     & \textBF{0.748} \\
    \bottomrule
  \end{tabular}
  \end{threeparttable}
\end{table}

\paragraph{Pseudolabel generation}

Instead of our proposed SAM regularization regime, which can handle overlapping SAM segments,
we can also use SAM masks more directly to generate pseudo-labels for semantic segmentation.
In this scenario, we utilize SAM as defined in \cref{sec:sam_preproc} to produce segments.
Now, an exponential moving average (EMA) version of the network under training is used to generate
semantic predictions on real data. For each object class, a best-fitting SAM segment is found through its IoU with the prediction,
thereby generating pseudo-labels for further training.
Since this training regime cannot be used from scratch due to instability of the generated pseudo-labels,
we apply it on a model pretrained on synthetic data for 300 epochs. The model is then trained in alternating
fashion on synthetic data and the pseudo-labeled real data.

In contrast to our SAM regularization, this pseudo-label generation method does not lead to
higher generalization ability compared to the baseline trained purely on synthetic data (see \cref{tab:ablations}).
We attribute this to situations where the model undersegments or oversegments objects in the real domain.
In this case, a pseudo-label-based approach is likely to select a SAM mask matching the under-/oversegmentation,
not yielding any improvement,
whereas our SAM regularization approach will lead the model to consider all possible segmentations found by SAM.

\paragraph{Wider domain gap}

\begin{figure}
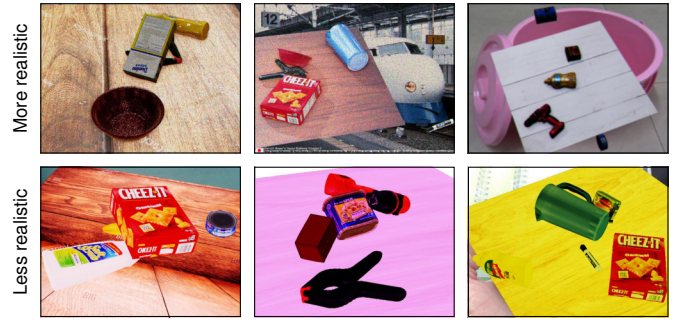

 \centering
 \begin{tikzpicture}[font=\sffamily\scriptsize]
  \matrix[matrix of nodes, inner sep=0pt, row sep=5pt, column sep=5pt, nodes={anchor=center}]{
    |[anchor=base,rotate=90]| More realistic &
    |[border on top]| \includegraphics[height=1.9cm]{images/unreal/real_syn1.jpg} &
    |[border on top]| \includegraphics[height=1.9cm]{images/unreal/real_syn2.jpg} &
    |[border on top]| \includegraphics[height=1.9cm]{images/unreal/real_syn3.jpg} \\
    |[anchor=base,rotate=90]| Less realistic &
    |[border on top]| \includegraphics[height=1.9cm]{images/unreal/unreal_syn1.jpg} &
    |[border on top]| \includegraphics[height=1.9cm]{images/unreal/unreal_syn2.jpg} &
    |[border on top]| \includegraphics[height=1.9cm]{images/unreal/unreal_syn3.jpg} \\
  };
 \end{tikzpicture}
 \vspace{-1ex}
 \caption{Widening the domain gap by reducing realism of the synthetic data generator.
 The bottom row has most available rendering options turned off.}
 \label{fig:unreal}
\end{figure}

What if source and target domains are further apart?
While domains with completely different styles are not our focus, we can artificially
widen the domain gap in a controlled manner by making the synthetic images less realistic.
In particular, we switch off the camera noise model, IBL light maps, shadows, and SSAO
in the Stillleben generator (see \citep{schwarz2020stillleben} for details). The resulting
images are clearly less realistic and less varied (see \cref{fig:unreal}) and are known
to lead to remarkably bad generalization performance~\citep{schwarz2020stillleben}.
\Cref{tab:ablations} shows results from training runs performed with this data,
with and without SAM regularization.
While the unregularized model clearly overfits on the less-varied source domain
data as expected, our proposed SAM regularization restores generalization.

\newlength\dl\setlength\dl{2cm}
\begin{figure}
 \centering
 \begin{tikzpicture}
  \matrix[matrix of nodes, inner sep=0, nodes={anchor=center,font=\sffamily\footnotesize}] {
    &[3pt] |[anchor=base]| Real/GT & |[anchor=base]| $D=1$ & |[anchor=base]| $D=2$ \\[2pt]
    |[anchor=base,rotate=90]| Input / Features &
    |[border on top]| \includegraphics[height=\dl]{images/d_comp/rgb/rgb001421.jpg} &
    |[border on top]| \includegraphics[height=\dl]{images/d_comp/d1/feat001421.png} &
    |[border on top]| \includegraphics[height=\dl]{images/d_comp/d3/feat001421.png} \\
    |[anchor=base,rotate=90]| Segmentation &
    |[border on top]| \includegraphics[height=\dl]{images/d_comp/labels/label001421.png} &
    |[border on top]| \includegraphics[height=\dl]{images/d_comp/d1/out001421.png} &
    |[border on top]| \includegraphics[height=\dl]{images/d_comp/d3/out001421.png} \\
  };
 \end{tikzpicture}
 \vspace{-1ex}
 \caption{Dense features learned with $D<3$. $D=1$ is visualized as grayscale,
 for $D=2$ the blue channel set to constant value.}
 \label{fig:d}
 \vspace{-1ex}
\end{figure}

\paragraph{The influence of $\alpha$ and $D$}

The loss scaling hyperparameter $\alpha$ was chosen to balance supervision and regularization losses.
As one can see in \cref{tab:ablations}, our model is not very sensitive to this parameter.

While choosing $D=3$ feature dimensions for the dense feature head is a convenient choice
for visualizing features, it is conceivable that higher-dimensional feature spaces might
lead to better separation of objects and grouping of parts.
However, as \cref{tab:ablations} shows, an increase to $D=12$ does not lead to better performance.
It is possible to increase $D$ further, but this requires optimization of the variance loss
computation due to memory constraints, e.g. by introducing random subsampling.
In the other direction, even for $D=1$ the model manages to properly separate segments in the resulting one-dimensional feature space (see \cref{fig:d}) and the SAM regularization scheme remains effective.

\subsection{Application in Robotic Setting}

\begin{figure}
 \centering

 \begin{tikzpicture}[font=\sffamily\scriptsize]
  \node[matrix,matrix of nodes,draw,rounded corners,nodes={anchor=center,inner sep=0pt},column sep=3pt,label={above,inner sep=1pt:Automatic training scene capture}] {
    \includegraphics[height=1.1cm]{images/eval/tiago_dataset/0001.png} &
    \includegraphics[height=1.1cm]{images/eval/tiago_dataset/0014.png} &
    \includegraphics[height=1.1cm]{images/eval/tiago_dataset/0021.png} &
    \includegraphics[height=1.1cm]{images/eval/tiago_dataset/0026.png} &
    $\dots$ &
    \includegraphics[height=1.1cm]{images/eval/tiago_dataset/0006.png} \\
  };
 \end{tikzpicture}
 
 \vspace{5pt}
 
 \begin{tikzpicture}[font=\sffamily\scriptsize]
  \matrix[eimgmatrix] {
   |[eimg]| \eimg{tiago_syn/real01.png} &
   |[eimg]| \eimg{tiago_syn/pred01.png} &
   |[eimg]| \eimg{tiago_samreg/pred01.png} &
   |[eimg]| \eimg{tiago_samreg/label01.png}
   \\
   |[eimg]| \eimg{tiago_syn/real02.png} &
   |[eimg]| \eimg{tiago_syn/pred02.png} &
   |[eimg]| \eimg{tiago_samreg/pred02.png} &
   |[eimg]| \eimg{tiago_samreg/label02.png}
   \\
   |[eimg]| \eimg{tiago_syn/real03.png} &
   |[eimg]| \eimg{tiago_syn/pred03.png} &
   |[eimg]| \eimg{tiago_samreg/pred03.png} &
   |[eimg]| \eimg{tiago_samreg/label03.png}
   \\
   |[eimg]| \eimg{tiago_syn/real07.png} &
   |[eimg]| \eimg{tiago_syn/pred07.png} &
   |[eimg]| \eimg{tiago_samreg/pred07.png} &
   |[eimg]| \eimg{tiago_samreg/label07.png}
   \\
   |[eimg]| \eimg{tiago_syn/real08.png} &
   |[eimg]| \eimg{tiago_syn/pred08.png} &
   |[eimg]| \eimg{tiago_samreg/pred08.png} &
   |[eimg]| \eimg{tiago_samreg/label08.png}
   \\[3pt]
   Test image & a) Synthetic-only & b) Full model & Ground truth \\
  };
 \end{tikzpicture}
 \caption{Qualitative segmentation results on a custom robotic test set.
 The unlabeled training data was automatically captured during autonomous grasping
 of operator-specified objects (top).
 The resulting dataset is used to SAM-regularize a model otherwise trained with
 labeled synthetic data (a), which gives improved results (b).
 }
 \label{fig:segresults_tiago}
\end{figure}

While demonstrating the usefulness of SAM regularization on YCB-Video and HomebrewedDB
is interesting, the question remains if it is actually useful in a robotic application.
For this purpose, we utilized a PAL Robotics TIAGo++ robot carrying an Orbbec Astra RGB-D camera.
Our TIAGo robot is used for research in household contexts and is capable of autonomously
grasping user-selected objects. During each such performed grasp, we automatically recorded
RGB frames showing the situation before, during, and after the grasp.
In this way, we recorded 28 scenes (i.e. object arrangements) with 3-21 frames each,
which we divide into 26 train and 2 test scenes.

Both, the quantitative results reported in \cref{tab:ablations} and the qualitative results shown in \cref{fig:segresults_tiago}, confirm that  SAM regularization is also effective in this setting.

\section{Conclusion}

We proposed a practical way to address the annotation bottleneck by using source domain data with annotations and target domain data without annotations, leveraging and learning from a foundation model.

Instead of focusing on improving the quality of the synthetic images, we make use of both domains to achieve a decrease in the Sim2Real gap.
Indeed, we have shown that a task-specific regularization using a related foundation model is possible and beneficial for the task itself.
Our method is universally applicable and we demonstrated its advantages for semantic segmentation on two well-known datasets. The evaluation of our method resulted in a mean IoU of 85\% and 79\% on YCB-Video and HomebrewedDB, a sufficient accuracy for real-world robotic tasks such as grasping.

The regularization term related to the SAM segments results in the learning of meaningful features, which
yields the observed improvement in task scores.
On YCB-Video, our model trained only on synthetic data and SAM regularization outperformed training on real data.
Therefore, we can regularize the model in a task-specific manner so that it performs more robustly in the target domain. 

Our method is not restricted to semantic segmentation. Joint training on a task on synthetic data and SAM segment regularization of real data is a generic tool that roboticists can use to reduce the effects of the Sim2Real gap.
We are convinced that our method will be applicable to other tasks, such as object detection, panoptic segmentation, 6D pose prediction,
 and other applications requiring dense features.

\section*{Acknowledgment}

\noindent{ \footnotesize
This research has been partially funded by the Federal Ministry of
Education and Research of Germany under grant no. 01IS22094A WestAI.
}

\printbibliography

\end{document}